\documentclass[12pt,a4paper]{article}
\usepackage[english]{babel}
\usepackage{graphicx}

\newcommand{\alt}{\mathbin{\lower 3pt\hbox
   {$\rlap{\raise 5pt\hbox{$\char'074$}}\mathchar"7218$}}}
\newcommand{\agt}{\mathbin{\lower 3pt\hbox
   {$\rlap{\raise 5pt\hbox{$\char'076$}}\mathchar"7218$}}}

\begin{document}
\setcounter{footnote}{0}
\setcounter{equation}{0}
\setcounter{figure}{0}
\setcounter{table}{0}
\vspace*{5mm}

\begin{center}
{\large\bf Can a Computer Laugh ? }

\vspace{4mm}
I. M. Suslov \\
P.N.Lebedev Physical Institute,\\
119334 Moscow, USSR\,\footnote{\,Present
address:

\noindent
P.L.Kapitza Institute for Physical Problems,
\\
119337 Moscow, Russia \\
E-mail: suslov@kapitza.ras.ru\\
} \vspace{4mm}
\end{center}

\begin{center}
\begin{minipage}{135mm}
{\large\bf Abstract } \\
A computer model of "a sense of humour" suggested previously
\cite{1,2}, relating the humorous effect with a specific
malfunction in  information processing,
is given in somewhat different exposition.
Psychological aspects of humour are elaborated more thoroughly.
The mechanism of laughter is formulated on the more general level.
Detailed discussion is presented for the higher levels of
information processing, which are responsible for a perception of
complex samples of humour. Development of a sense of humour in
the process of evolution is discussed.
\end{minipage}
 \end{center}
 \vspace{5mm}


\begin{center}
{\bf 1. Introduction}
\end{center}
\vspace{3mm}

In recent decades we observe a swift penetration of computers
in different areas of human life: computers play chess, compose
music, prove theorems and so on. However, there is one area,
which  is considered as unreachable for computers: it is an area
of emotions. Already the thought that a computer can "feel"
is often considered as
sacrilegeous.
Are possibilities of computers really so restricted?

Below we suggest an analysis, which shows the possibility to
endow computers by simplest human emotions and in particular
a "sense of humour", i.\,e. ability to react on jokes.
The humorous effect is
treated as a specific malfunction in the processing of information,
conditioned by the necessity of a quick deletion from consciousness
of a false version.
The suggested model leads to a natural resolution for the problem
of the biological function of humour. Indeed, what
a purpose was followed by nature, when it has provided us with
a sense of humour?   The bare fact that there
exists a complex biological mechanism which causes specific
muscular contractions (laughter) as a reaction to a definite
combination of sound or visual images leads us to conclude that
the sense of humour originated at early stages of the
evolution\,\footnote{\,According to Darwin \cite{3},
antropoid monkeys  possess a clearly distincted sense of humour.}
and is not a product of human civilization. It will be shown
below, that endowing a biological object by a sense of humour
leads to quickening the transmission of processed information
into consciousness and to a more effective use of brain
resources.

The paper is organized as follows. The psychological analysis of
jokes based on Freud's classification \cite{4} leads to a
hypothesis on the conditions of arising the humorous effect
(Sec.\,2).  The natural model for the primary processing of
information (Sec.\,3) shows that the humorous effect, as other
simple emotions (Sec.\,4), arises already at this level (Sec.\,5).
Interpretation of more complex samples of humour requires
investigation of the higher levels of information processing
(Sec.\,8\,--\,10). The old idea by Spencer \cite{5} on relation of
laughter with release of nervous energy acquires a clear physical
sense, if information on the properties of neural networks
\cite{6} is taken into account (Sec.\,6). We shall see in Sec.\,7
that the model gives natural explanation to well-known facts:
different succeptibility of people to humour, absense of humorous
effect from a trite joke, nervous laughter, etc. Development of
the sense of humour in the process of evolution is discussed in
Sec.\,11. At last, in Sec.\,12 we discuss a question
carried out to a title of the paper.

\vspace{6mm}
\begin{center}
{\bf 2. Humour from the psychological viewpoint}
\end{center}
\vspace{3mm}

The technical ways of creating jokes was classified by
Freud in the book \cite{4}. Let us give an example to each
technical method, in order to obtain the representative selection
from enormous number of jokes\,\footnote{\,All examples are taken
from \cite{4}, but given more briefly, if possible. Another
classification of technical aspects of wit is suggested in
\cite{7}.}.

\vspace{3mm}
\noindent
1. Condensation:

(1*) {\it\quad\,\,  Rotshild treated me famillionairely. }

\vspace{3mm}
\noindent
2. The repeated use of the same material:

(2*) {\it\quad\,\,  Put not your trust in money, but your
money in trust. }

\vspace{3mm}
\noindent
3. Ambiguity (a play on words, double entendre, etc):

 (3*)  {\it \quad\,\, During Dreifus's process:

\qquad\qquad\,"This girl is like Dreifus:
army does not believe in her innocence". }

\vspace{3mm}
\noindent
4. Pun:

 (4*)  {\it \quad\,\, I have been in Calcutta as well, as this
 "Calcutta roast".}

\vspace{3mm}
\noindent
5. Displacement (a shift of the accent from one part of the
phrase to another):

 (5*)  {\it \quad\,\, The horse tradesman: "If you mount this
 horse at 4 in the morning

\qquad\qquad\, then at 7 in the morning you will be at
Pittsburg."

\qquad\qquad The customer: "But what shall I do in Pittsburg at 7
in the morning?"}

\vspace{3mm}

\noindent
6. The use of mistakes in the reasoning.
\vspace{1mm}

\noindent
(a) A sense in the nonsense:

 (6*)  {\it \quad\,\, Isaak was a bad soldier. The officer
 annoyed by him advised:

\qquad\qquad "Listen, Isaak, buy your own gun and work
separately."}

\vspace{3mm}

\noindent
(b) Sophistical mistakes:

(7*) {\it\quad\,\, Is this a place where Duke of Wellington said
his famous words?

\qquad\qquad Yes, it is the same place but he never said such
words.}

\vspace{3mm}

\noindent
(c) Automatic mistakes:

(8*) {\it\quad\,\, The marriage agent in the home of bride:

\qquad\qquad --- You can judge by these things, how rich are
these people.

\qquad\qquad --- But cannot they borrow these things to produce
an impression?

\qquad\qquad --- Oh, nuts! Who will trust even a
thing to these people?}

\vspace{3mm}

\noindent
7. Unification (finding the common in the different):

(9*) {\it\quad\,\, Prince, travelling over his estates, notes
a man very much alike

\qquad\qquad\,  his own high person:

\qquad\qquad --- Did not your mother serve in the manor house?

\qquad\qquad --- Mother -- no, but father -- yes.}

\vspace{3mm}

\noindent
8. Indirect representation.
\vspace{1mm}

\noindent
(a) Representation by the opposite:

(10*) {\it\quad\,\, --- Can you conjure up the spirits?

\qquad\qquad\, --- Yes, I can. But they never come.}

\vspace{3mm}
\noindent
(b) Representation by the exaggeration:

(11*) {\it\quad\,\, --- Does she dye her hairs?

\qquad\qquad\, --- No, they were always such. Since the time she
has bought them.}

\vspace{3mm}

\noindent
(c) Representation by the similar (a hint):

(12*) {\it\quad\,\, Prince N. is a complete
idealist.}

\vspace{3mm}

\noindent
(d) Representation by the comparison:

(13*) {\it\quad\,\, A single-sleeping church chair.}

\vspace{3mm}
\noindent
Analysing these examples, one can come to a following hypothesis:
the humorous effect is related with interference in the human
consciousness of two mutually exclusive images (versions,
estimates). The statement of such kind (the so called concept of
incogruity) was advanced by the Scotch poet Beattie
in l776 \cite{8} and is admitted in some form by  most of
researches (see e.\,g. \cite{7,9,10,11}).

The given jokes are constructed according two principal schemes:
the "switching scheme" and "ambiguity scheme". For the jokes of
the first group, two mutually exclusive versions follow in
a definite succession: firstly, one of images arises, and then
it is "switched" to another image. Thus,
the words of the tradesman in example (5*) realized
as "the characteristic of horse speed" take on the
interpretation "giving directions
how to reach Pittsburg by 7 in the morning".
The officer's advise in example (6*) is looking as "nonsense",
but later is realized as "a deep penetration to Isaak's dreams"
(e.\,g. to buy a shop and work separately). In example (7*),
the second remark at first gives the
impression of being "natural" or "logical" but later is perceived
as "absurd". Analogously,  a "natural" answer of
the agent in example (8*) is realized as "silly" and going apart with his
purposes. In example (9*), the prince's version "you are a
bastard" is turned to himself. The answer "yes" in example (10*)
in fact appears to be "no". In example (11*), a version "her
hairs are so good that they need no drying" is switched by
version "she has no hairs".

The jokes of the second group, constructed in accordance with
the "ambiguity scheme", are characterized by equality
of two versions: they exist more or less simultaneously and one
cannot establish their correct succession. In example (1*),
there is interference between "familiarly" and "millionaire".
In example (2*), the word "trust" in the second part of the
phrase can be interpreted as "reliable place" (in accordance
with its immediate sense) or as  "confidence" (in the analogy
with the first part).  The word "innocence" in
example (3*) can be interpreted as
"guiltlessness" or
"virginity".  In example (4*), "Calcutta roast" (the name of the
dish) is interfered with "roast from Calcutta".
In example (12*), "complete
idealist" is resembling "complete idiot", while example (13*)
deals with a chair for "sitting" or "sleeping" in the church.

The humorous effect is caused not only by a "wit" discussed
above, but also by the "comic", the main characteristic
of which is "deviation from the norm". Examples of the comic are
exaggerated movements of a clown, grimaces, naive saying
of children, accidently said silliness or nonsense, etc.
The comic is created by  caricature, parody, imitation,
disguise, exposure, unmasking, and and so on.
The humorous effect from the comic agrees with the formulated
hypothesis, if one accept (see e.\,g. \cite{4}) that there are
oscillations between two contrasting notions, the observed
comic and realized "norm".

The laughter from tickling can be connected with the attempt
of the brain to localize the place of irritation of skin; the
result of such localization is invariably rejected because the
irritated place is changed unpredictably: that is the reason
why the tickling should be done by another person.

\vspace{6mm}
\begin{center}
{\bf 3. The primary processing of information  }
\end{center}
\vspace{3mm}

We begin the formulation of the computer model of "a sense of
humour" from the consideration of the primary processing
of information. Suppose that a succession of symbols $A_1$,
$A_2$, $A_3,\ldots$ ("text") is introduced from the outside world
to the brain: it can be a succession of words during visual or
auditory percepption.  In the brain a set of images $\{B_n\}$ is
associated with each symbol $A_n$: for example a set of meanings
(a dictionary family) is put in correspondence to each word. The
problem of the primary processing of information consists in
choosing one image $B_n^{i_n}$ (which is implied in the given
context) from the set $\{B_n\}$. The text will be considered as
"understood" if the succession $B_1^{i_1}$, $B_2^{i_2}$,
$B_3^{i_3},\ldots$ (which visually can be imagined as the
trajectory in the space of images --- see Fig.\,1) is put in
\begin{figure}
\centerline{\includegraphics[width=5.1 in]{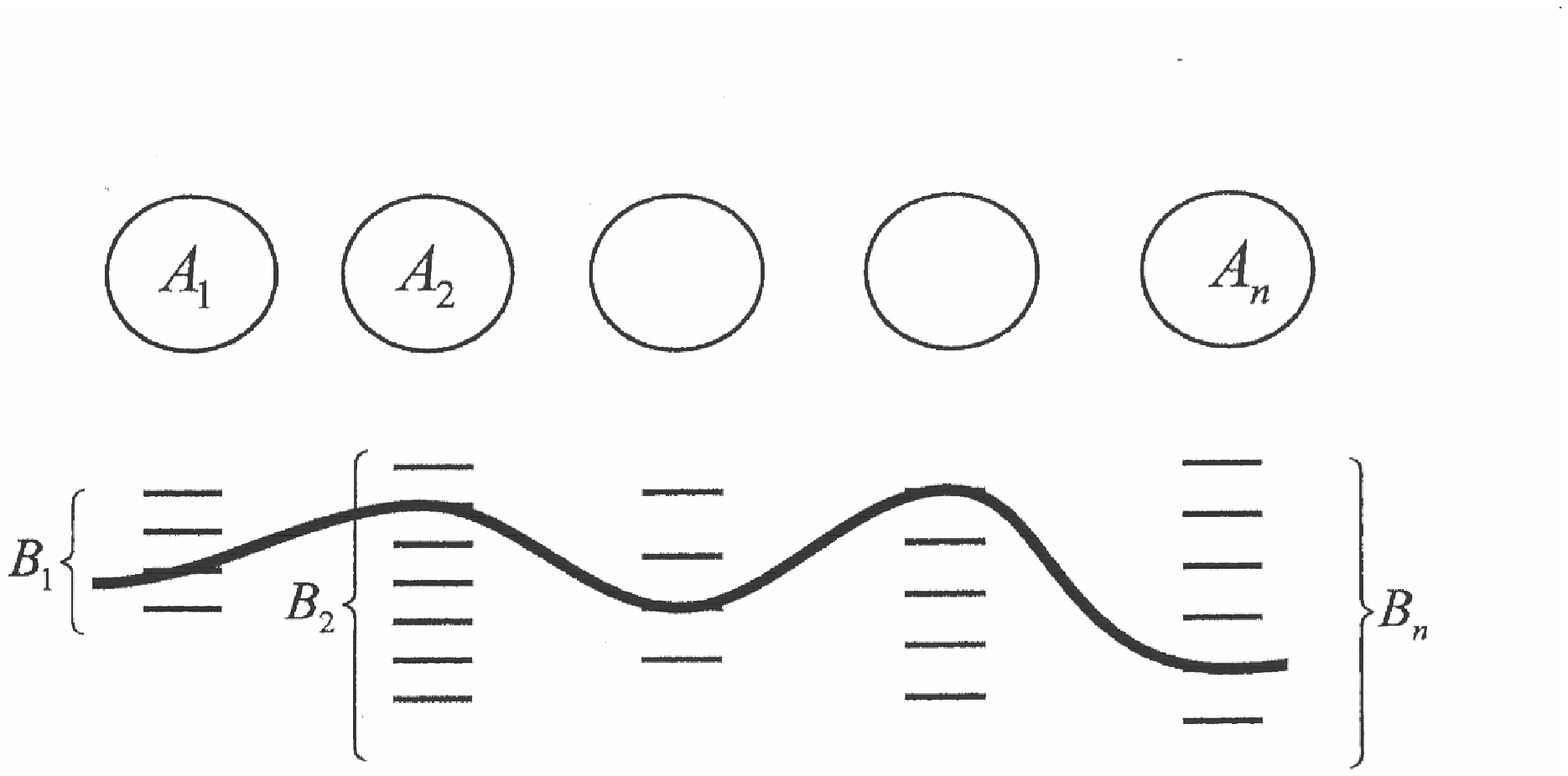}}
\caption{The scheme of information processing:
a set of images $\{B_n\}$ is put in correspondence to  each
symbol $A_n$ and  one image $B_n^{i_n}$ should be chosen
from the set $\{B_n\}$. The succession
$B_1^{i_1}$, $B_2^{i_2}$, $B_3^{i_3},\ldots$
looks as a  "trajectory" in the space of images. } \label{fig1}
\end{figure}
correspondence to the succession $A_1$, $A_2$, $A_3,\ldots$ In
principle, the algorithm of the primary processing of information
consists in the following:

(1) all possible trajectories in the image space are constructed;

(2) a certain probability is ascribed to each trajectory on the
basis of the information on the correlation of images stored in
memory;

(3) the most probable trajectory is chosen.

Only step 2 is nontrivial here, i.\,e. the algorithm of
ascribing the probability to a given trajectory. For example,
such algorithm can be based on the binary correlations of images;
in this case the set $p_{ij}$ should be stored in the memory
where $p_{ij}$ is the probability of the event that in a
meaningful text image $i$ will be followed by image $j$; the
probability of a trajectory $ijkl\ldots$ is given by the product
$p_{ij} p_{jk} p_{kl} \ldots$. The probabilities
$p_{ij}$ can be obtained by the statistical treatment in the
course of the "learning" process, during which a
sufficiently long fragment of the "deciphered" text (i.\,e.
recorded in images but not symbols) is introduced to the brain.
A more complex algorithm can take into account the correlation
between $n$ images ($n>2$): then the probabilities
$p_{i_1 \ldots i_{n-1};\, i_n}$ of the succession of images
$i_1 \ldots i_{n-1}$ followed by image  $i_n$ should be
stored in the memory. It is possible to base the algorithm on
binary correlations but with the syntactical connections
taken into account\,\footnote{\,The syntactic structure of a
sentence may be represented in the form of "a tree", so that
each dependable word is linked with its "host". The probability
of a trajectory is represented as a product of binary
probabilities according to the structure of a syntactical "tree".
The practice of machine translation \cite{12,13,14} shows that
the syntatic structure in most cases can be established by purely
grammatical analysis (word order, adherence to a part of speach,
harmonization of the endings, etc.) and can be considered
as known for the purposes of the present  work.  }
and so on.
Algorithms of such type are being worked out in the investigations on
machine translation \cite{12,13,14}; the concrete form of the
algorithm is not essential for the following.

The number of operations demanded for the realization of
any algorithm of such kind increases exponentially with the
length of the text. So only fragments of the text containing no
more than $L$ symbols can be immediately treated by such a
method. How can longer texts be processed? The natural
possibility is the following: during the processing of the first
$L$ symbols not one but several ($M$) of the most probable
trajectories are remembered; then translation on one step is made
--- the fragment from the second to the $(L+1)$-th symbols is
considered --- and for each of the $M$ conserved trajectories all
possible continuations are constructed; then again $M$ of the
most probable trajectories are conserved and so on. It is
reasonable to make the number $M$ variable, so at each stage as
many trajectories are remembered as the operative memory can
hold. In the whole, the process looks as follows (Fig.\,2):
\begin{figure}
\centerline{\includegraphics[width=5.1 in]{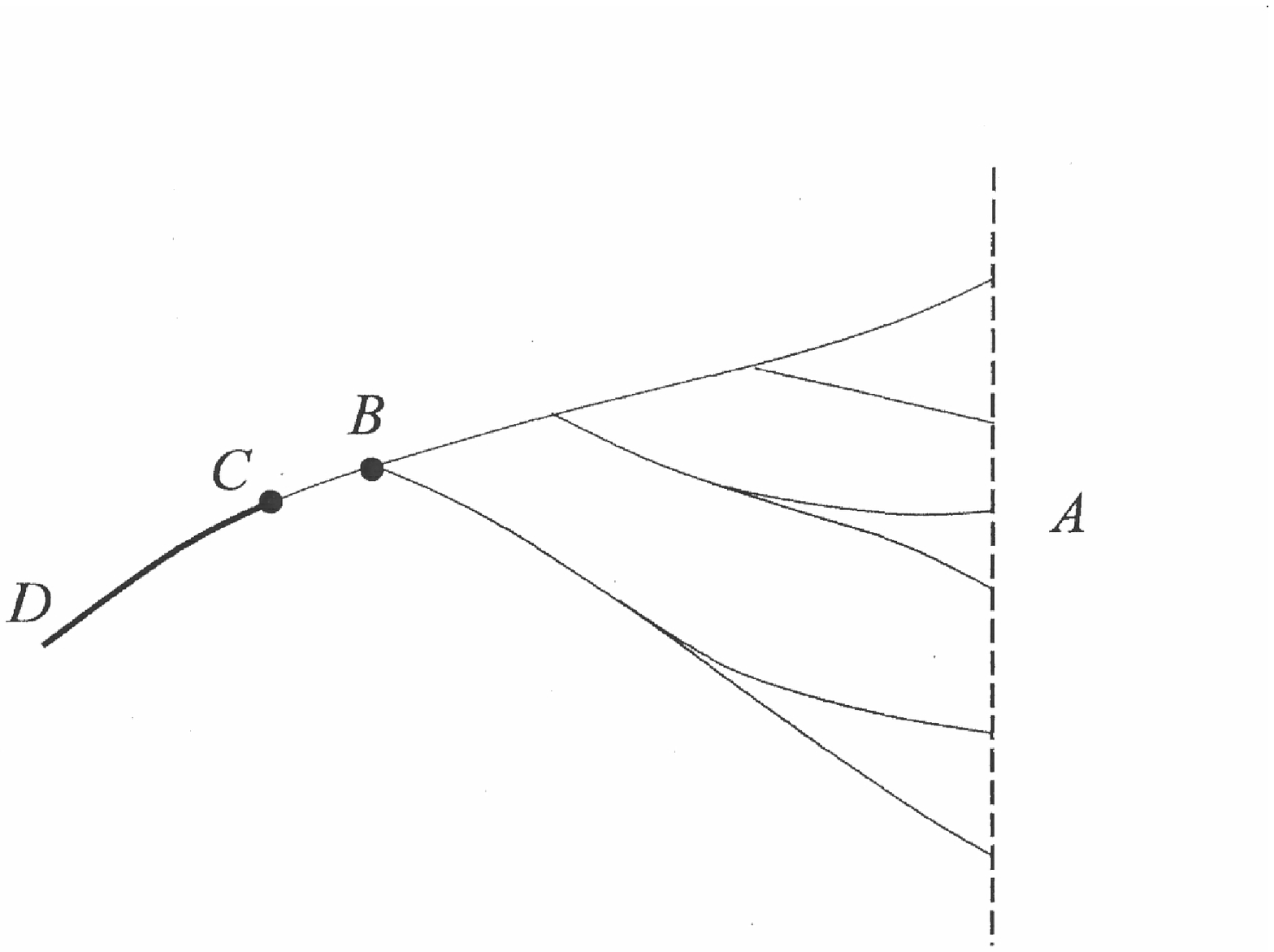}}
\caption{The visual imagination of information processing:
thin solid lines are trajectories conserved in
operative memory, $A$ is a front, $B$ is the point where the
branching is over, $CD$ is a fragment of deciphered trajectory
transmitted to consciousness. } \label{fig2}
\end{figure}
immediately after the front $A$ the trajectory is branched
heavily; at a certain point $B$ the branching is over (the
distance between $A$ and $B$ is restricted by the volume of
operative memory provided for remembering the trajectories); the
deciphered part of the trajectory $DC$ with some delay $AC$ is
transmitted to the consciousness of the man and is realized by
him as a thought (while the whole process takes place in the
subconscious and is not perceived immediately).

\vspace{6mm}
\begin{center}
{\bf 4. The role of emotions in information processing  }
\end{center}
\vspace{3mm}

If numbers $L$ and $M$ are sufficiently large and the algorithm
of calculating the probability of $N$-symbol trajectory is
good enough, then the described scheme will operate successfully.
However the probabilitistic nature of the algorithm makes
mistakes inevitable: so a mechanism is desirable for minimizing
their consequences. Such mechanism exists and it consists in
communicating to the consciousness some information
about the course of the processing in the subconsciousness; the
man perceives such information as emotions.

For example, such parameters of the process are essential
as the probability $p_{max}$ of the trajectory transmitted
to consciousness and the probability $p_{comp}$ of the most
probable of the competing trajectories.  The high values
of $p_{max}$ and  $p_{max}/p_{comp}$ signal a successful
course of the process and are perceived as positive emotions
(pleasure, confidence): the information obtained is considered
as reliable. The low values of $p_{max}$ and  $p_{max}/p_{comp}$
signal an unsatisfactory course of the
process and  are realized as negative emotions (annoyance, doubt):
the corresponding information should not be taken too seriously.
For very low values of $p_{max}$  no versions are transmitted to
consciousness (complete incomprehension) and so on.

The possible relationship of emotions with the parameters
of the process can be illustrated on the basis of the
semi-empiric "emotion formula" proposed by Simonov [10]
$$
{\cal E}={\cal N} (I-I_0)
$$
where $\cal E$ is the emotion strength (which is objectively
measured by the pulse rate, the blood pressure etc.), $\cal N$ is
a strength of some need, $I_0$ is the quantity of information
demanded for the satisfaction of this need, $I$ is the quantity
of information the subject has at his disposal (both informations
are estimated subjectively). An emotion is positive (${\cal
E}>0$) for $I>I_0$ and negative for $I < I_0$.  We can suppose
that in the course of processing $\cal N$ is the
need in information and the different parameters of the process
determine $I$ and $I_0$ for different emotions. For example,
$p_{max}$  can be used as $I$ for the emotion "pleasure of
understanding - annoyance of incomprehension" (accordingly,
$I_0$ is the typical value of $p_{max}$ ensuring the satisfactory
course of the process).  Analogously, $p_{max}/p_{comp}$ can be
used as $I$  if $\cal E$ is the emotion "confidence\,--\,doubt"
and so on.

These speculations lead us to conclude that the emotion
expressing the humorous effect is also related to some specific
situation in the processing of information.

\vspace{6mm}
\begin{center}
{\bf 5. The humorous effect   }
\end{center}
\vspace{3mm}

Let us discuss the nature of the delay of point $C$ with
respect to front $A$ (Fig.\,2). At first sight, point $C$ in a
reasonably organized system should be always behind point
$B$ or   coincide with it: it is just the variant we
surely choose writing the computer program. But for a human
as well as for any biological object such a variant is completely
unsatisfactory. The matter is that the delay of point $C$
 with respect to front $A$ results in the time interval
$\tau_{AC}$  during which the information introduced to the
brain does not appear in the consciousness (the man sees a
bear but he is not aware of this). The prolongation of the
interval $AC$ is  obviously dangerous while the
interval $AB$ can drag out for objective reasons (the man cannot
decide what he sees: a bear or a bush shaped like a bear). So
the interval  $AC$ should have the upper bound $\tau_{max}$
on the time scale: if time delay $\tau_{AB}$ corresponding to
the interval $AB$ is less than $\tau_{max}$ then point
$C$ coincides with point  $B$ (Fig.\,3,a);  if
\begin{figure}
\centerline{\includegraphics[width=5.1 in]{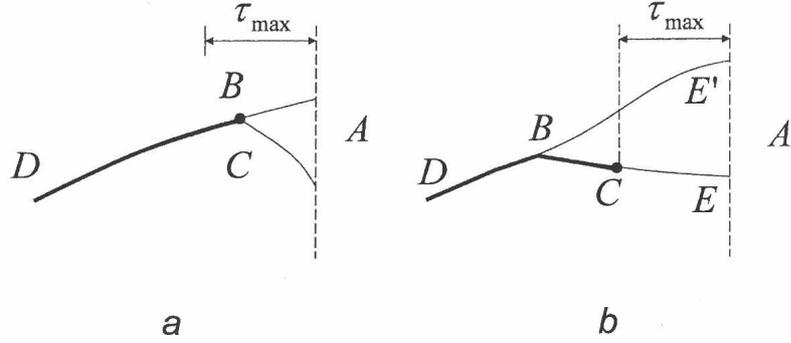}}
\caption{The parameter $\tau_{max}$  is the upper bound of the
time interval corresponding to delay of point $C$ with respect
to front $A$; (a) $\tau_{AB}<\tau_{max}$, (b)
$\tau_{AB}>\tau_{max}$. } \label{fig3}
\end{figure}
$\tau_{AB}>\tau_{max}$, then $\tau_{AC}=\tau_{max}$
and point  $C$ outruns point $B$ (Fig.\,3,b). In the latter
case, the most probable version $DE$ is transmitted to the
consciousness while competing versions ($DE'$) are conserved in
the operative  memory (Fig.\,3,b) --- their deletion is unreasonable
because the brain has resources to continue the analysis.
If in the course of the subsequent movement of front $A$
the trajectory $DE$ continues to
have the maximum probability, then the competing trajectory $DE'$
will be deleted and the time will be saved as a result. If in the
course of the movement of front $A$ the probability of $DE$
falls below the probability of $DE'$, then the brain will have
a possibility to correct the mistake. In this case, however,
the specific malfunction occurs: the fragment
$BC$ transmitted to consciousness should be immediately deleted
and replaced by the fragment of trajectory $BE'$.
Psychologically this malfunction is perceived as interference
of two incompatible versions: version $BC$ fixed by the
long-term memory and the newly appeared version $BE'$. The
described specific malfunction can be identified with
"a humorous effect".

Indeed, the situation described is exactly reproduced in
the course of the interpretation of jocular expressions. For
example,  in the joke
\vspace{3mm}

(14*) {\it\quad\,\,  "My Uncle William has a new cedar chest."

\qquad\qquad "So! Last time I saw him he just had a wooden leg." }

\vspace{3mm}
\noindent
two incompatible versions arise in the
subconsciousness during the analysis of the first remark: in
the first of them ($DE$) the word "chest" is treated as "box"
while in the second ($DE'$) it is treated as "breast". In the
context of the given sentence version $DE$ ("box") is more
probable and is transmitted to consciousness. The appearance of the
word "leg" in the second remark makes version $DE$ less probable
and increases the probability of version $DE'$ ("breast"): this
gives rise to a humorous effect.

It is essential to emphasize that the existence of a humorous
effect is not to any degree unavoidable: nature had a possibility
to avoid it in one of the two manners:
(1) by delaying the transmission of trajectory $DE$ to consciousness
till trajectory $DE'$ is naturally discarded, or (2) by quickening
the transmission of $DE$ by deleting $DE'$ simultaneously. However,
in the first case the time the information reaches consciousness
is delayed and in the second case the brain resources are not
completely used: so nature resolves this problem at the cost of
psychological confusion.

In the process of evolution the optimal value of  $\tau_{max}$
is achieved which ensures the compromise between the reliability
of information and the speed of its obtaining (people with long
$\tau_{max}$ will be eaten by a bear, while people with short
$\tau_{max}$  will confuse every bush with a bear and will be
incapable of getting food). For the optimal value of $\tau_{max}$
the inequality $\tau_{AB}<\tau_{max}$
is satisfied as a rule, and a humorous effect is rare enough
in the natural conditions; but it can be easily produced by
specially constructed witticisms and comics.

\vspace{6mm}
\begin{center}
{\bf 6. The mechanism of laughter   }
\end{center}
\vspace{3mm}

According to Hopfield model \cite{6}, a neural network can
be described (in a reasonable approximation) as a system of
interacting spins. Two states of a neuron ("excited" and "rest")
can be corresponded with two states of a spin ("spin up" and
"spin down"); describing these states by Ising variables
$\sigma_i=\pm 1$  (spins or neurons are
numbering by index $i$), one can introduce an  energy $E$ of the
spin system:
$$
E=\sum\limits_{ij} J_{ij} \sigma_i \sigma_j
$$
where $J_{ij}$ are exchange integrals (for the spin system), or
the links between neurons (for a neural network). It is possible
to introduce also the "temperature" $T$, "magnetic field" $H$,
and a rate of dissipation $\gamma$ ($\gamma^{-1}$ is a
characteristic time of relaxation to a
stable state)\,\footnote{\,For a neural network, $T$ is
determined by the noise level in it, $H$ can be created by a
special group of neurons ("magnet"), while $\gamma^{-1}$ is a
relaxation time to one of "written images" (the latter
can be identified with stable states in the corresponded "spin
glass"). }.

According to Sec.\,2, the humorous effect arises in a situation,
when some information should be quickly deleted from the
consciousness. On the physical language it corresponds to
necessity of quick "nullification" of some portion of neural
network: all neurons, which become excited in the course of
writing certain information, should be transfered to the rest
state.  Such "nullification" can be carried out by switching a
strong magnetic field, which makes a state "spin up"
energetically unfavorable: so after a time $\sim\gamma^{-1}$ all
spins will come into the state "spin down". It is possible to
delete information for a time lesser than $\gamma^{-1}$, if  (in
the presence of the magnetic field) one connect a nullified
region with a "reservoir", i.\,e. large portion of neural network
in the rest state. Then an access of neuron energy is throwing
out to reservoir. The role of reservoir can be played by portions
of neural network, unrelated with a function of thinking (in
order not to arise parasitic thoughts), i.\,e.  motor cortex. In
this case, deleting of wrong version from consciousness arises
contraction of certain muscles, i.\,e. laughter.

It is easy to notice the similarity with old idea by Spencer
\cite{5} that the humorous effect is caused by a release of
nervous energy which is transformed into muscular contractions.
This idea was suppported by Darwin \cite{3} and Freud \cite{4}
but was criticized by subsequent investigators \cite{16} on the
grounds that the concept of "nervous energy" was difficult to
define. In fact,  it is possible to define
the  "energy" for neural networks only under condition
$J_{ij}=J_{ji}$, which is not very realistic for actual neurons;
so the concept of "nervous energy" should not be taken too
seriously. Nevertheless, the qualitative picture following
from the Hopfield model looks rather reasonable, and
appears in good correspondence with Spencer's hypothesis:
it looks as if the excitation energy of neurons is thrown out
into the motor system.

The release of the nervous energy in the humorous situation
was argued by Spencer on the basis of the concept of decreasing
incongruity --- the transition from the high style (state with
rich associations) to the low style (state with poor
associations). Such treatment of humorous effect is clearly
restricted and cannot  claim universality. In our scheme
"the release of  nervous energy" (in the conditional sense
discussed above) is connected with the necessity of deleting
the false version from  consciousness.

Since laughter is treated as an unconditioned reflex to the
humorous effect, then an explanation should be given to the
well-known exclusions, when laughter is forced out by secondary
emotions. For example, laughter can be forced out by indignation
(an indecent anecdote was told to a puritan),  fear (a bush
suddenly appeared to be a bear), pity (a man in front of you has
slipped on a melon skin and hurt himself seriously), shame (you
have slipped on a melon skin) etc. Spencer's hypothesis accounts
for these cases, if we suppose that "released nervous
energy" is transmitted not to motor cortex but to other regions
of the nervous system (the limbic system) where it is used to
excite secondary emotions. The fact, that a joke produces
a maximum effect if it is
expressed extremely laconically, has an analogous explanation
\cite{4}:  the laconism reduces the possibility of side
associations that can absorb the "nervous energy".

A man can regulate the level of muscular reaction by
transmitting the excitation of neurons to different regions
of the motor cortex: this can explain the dependence of
laughter on the mood,
the psychological motivation, the existence of laughing
environment and so on  \cite{17}.

\vspace{6mm}
\begin{center}
{\bf 7. Some consequences  }
\end{center}
\vspace{3mm}

The model described accounts for a number of well-known
facts.

\vspace{2mm}

{\it Absense of a humorous effect from a hackneyed
joke}
is a consequence of the fact that a man knows of the
existence of two incompatible versions beforehand and avoids the
transmission of the clearly false version to consciousness (for
example, knowing that in joke (14*) the "chest" turns out to be a
"breast" he is not tempted to interpret it as "box").

\vspace{2mm}

{\it The role of intonation in telling jokes} is connected mainly
with timing (tempo, arrangement and duration of pauses,
etc), which can be taken into account by incorporating an
appropriate number of "gaps" in succession $A_n$. The quick
pace of telling does not give time for the false
version to be transmitted to consciousness and interval $BC$
(Fig.\,3) turns out small or absent.  The slow pace of telling
increases the lengths of trajectories due to "gaps" and the
competing trajectory $BE'$ (Fig.\,3) is deleted from the operative
memory; so the swithing of versions becomes impossible.

\vspace{2mm}

{\it Different susceptibility of people to humour}\,\footnote{\,We
have in mind the principal susceptibility to
humour without regard to the cases when individual peculiarities
give rise to inadequate reaction to a concrete joke. The examples
are failure to understand a joke due to the absence in memory
of a necessary image, peculiar view of the "norm" while perceiving
the comic, the forcing out of laughter by side emotions (Sec.\,6)
 and so on.}
is connected
(in case of the equal intellectual level) with the differences in
the delay $\tau_{max}$. People with large  $\tau_{max}$
seldom laugh because point $C$ seldom outruns point $B$. Conversely,
people with small $\tau_{max}$ are aware of a humorous effect
even in cases that most people do not see as funny. Supposedly,
 $\tau_{max}$  is diminished by alcohol and this is
a cause of the unmotivated  gaiety.
At fixed $\tau_{max}$ the susceptibility to humour correlates with
the volume of the operative memory, which determines the average
length of the interval $AB$ (Fig.\,2).

\vspace{2mm}

{\it Nervous laughter}. If a variety of unpleasant impressions
rushes at a man and there is danger of the overstrain of the
nervous system then the organism forcibly deletes the unpleasant
information and replaces it by neutral: this gives rise
to the reflectory laughter.

\vspace{6mm}
\begin{center}
{\bf 8. The higher levels of the information processing  }
\end{center}
\vspace{3mm}

In principle, the proposed scheme explains all types of
humour considered above if we do not specify the nature of the
symbols $A_n$.  However, if practical modelling of the sense of
humour is intended, then it is convenient to consider symbols
$A_n$ as usual words. In this case the preceeding consideration
gives us an explanation only for jokes based on the switching
of meanings of a separate word (example 14*). In order to explain
more complex types of jokes it is necessary to consider the
higher levels of the information processing.

At the first level of processing the succession of primary
images is put in correspondence to a succession of symbols $A_n$.
Suppose that besides the first level a second level exists where
the succession of larger (secondary) images is put in correspondence
to the succession of primary images. On the third level
even larger (tertiary) images are constructed and so on (Fig.\,4).
\begin{figure}
\centerline{\includegraphics[width=5.1 in]{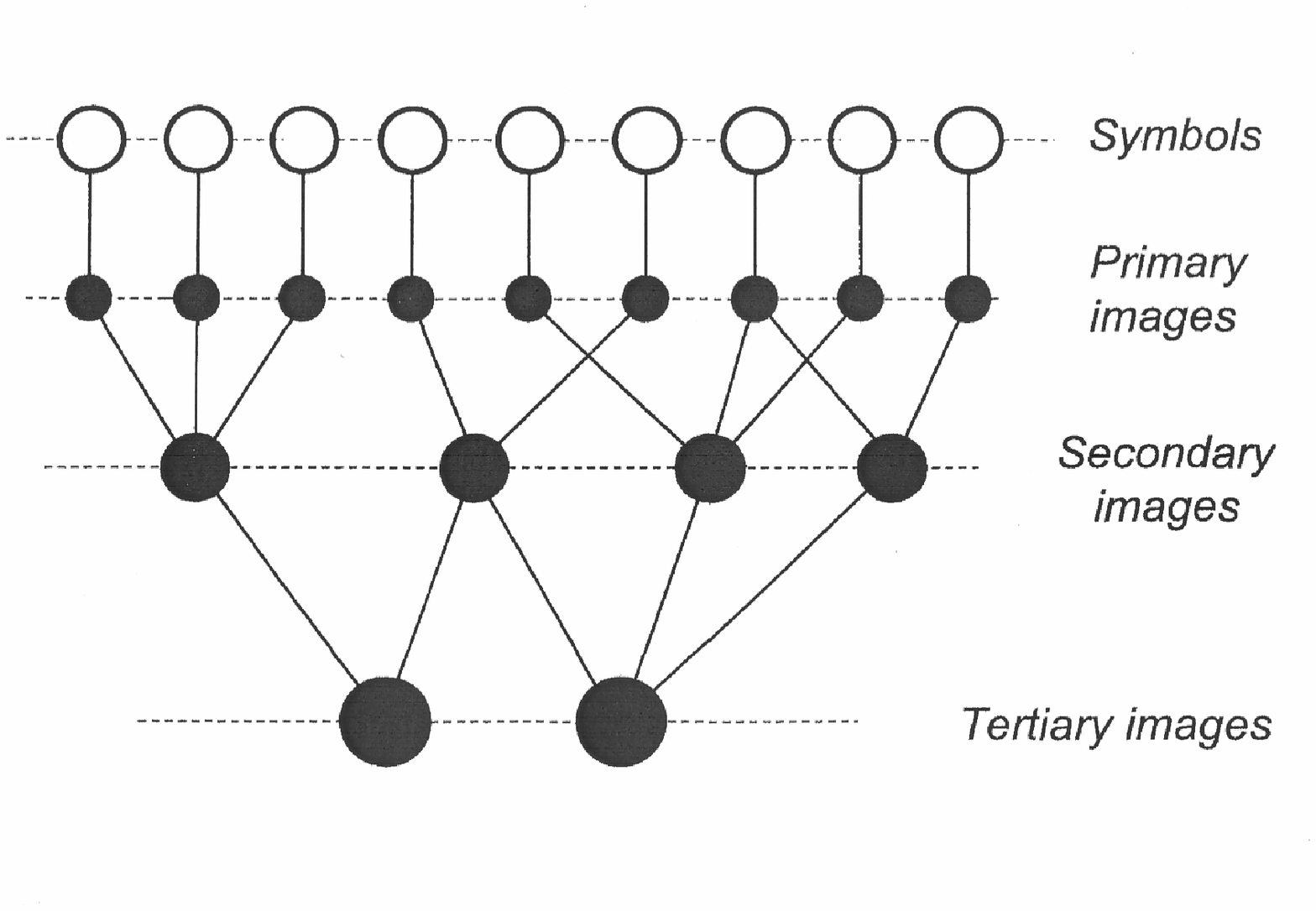}}
\caption{Higher levels of information processing.} \label{fig4}
\end{figure}
The obtained images are transmitted to consciousness in such
a way that images of each level go to a special channel. The
whole information is perceived on several levels simultaneously
("stereoscopically").
The principal algorithm of transition from level $k$ to level
$(k+1)$ consists in the following. Each image stored in memory
(e.\,g. "cow") is associatively connected with the smaller
images ("horns", "hoofs") as well as the larger ("herd") and
the more   abstract   ("animal")   images.   Suppose $A_n$ is
a primary image  and  the   set $\{B_n\}$    of images   associated
with  $A_n$ contains  only  larger or   more   abstract   images
than  $A_n$.  Then the problem  of constructing the secondary
version consists in choosing the appropriate image $B_n^{i_n}$
and is analogous to the problem of the
primary processing\,\footnote{\,The secondary version does not
necessarily differ from the primary one by enlarging all images
without exception; so in real algorithm it is reasonable to
include all images associated with $A_n$ into the set $\{B_n\}$
but at the same time to introduce the weighting function which
would increase the probability of versions with larger images.}.
If in a primary
succession $A_n$ several images are contained (e.\,g. "horns",
"hoofs", "tail"), which are parts of a larger image ("cow") then
the latter will be contained in each of the corresponding sets
$\{B_n\}$ (Fig.\,5); the most connected version will contain
\begin{figure}
\centerline{\includegraphics[width=5.1 in]{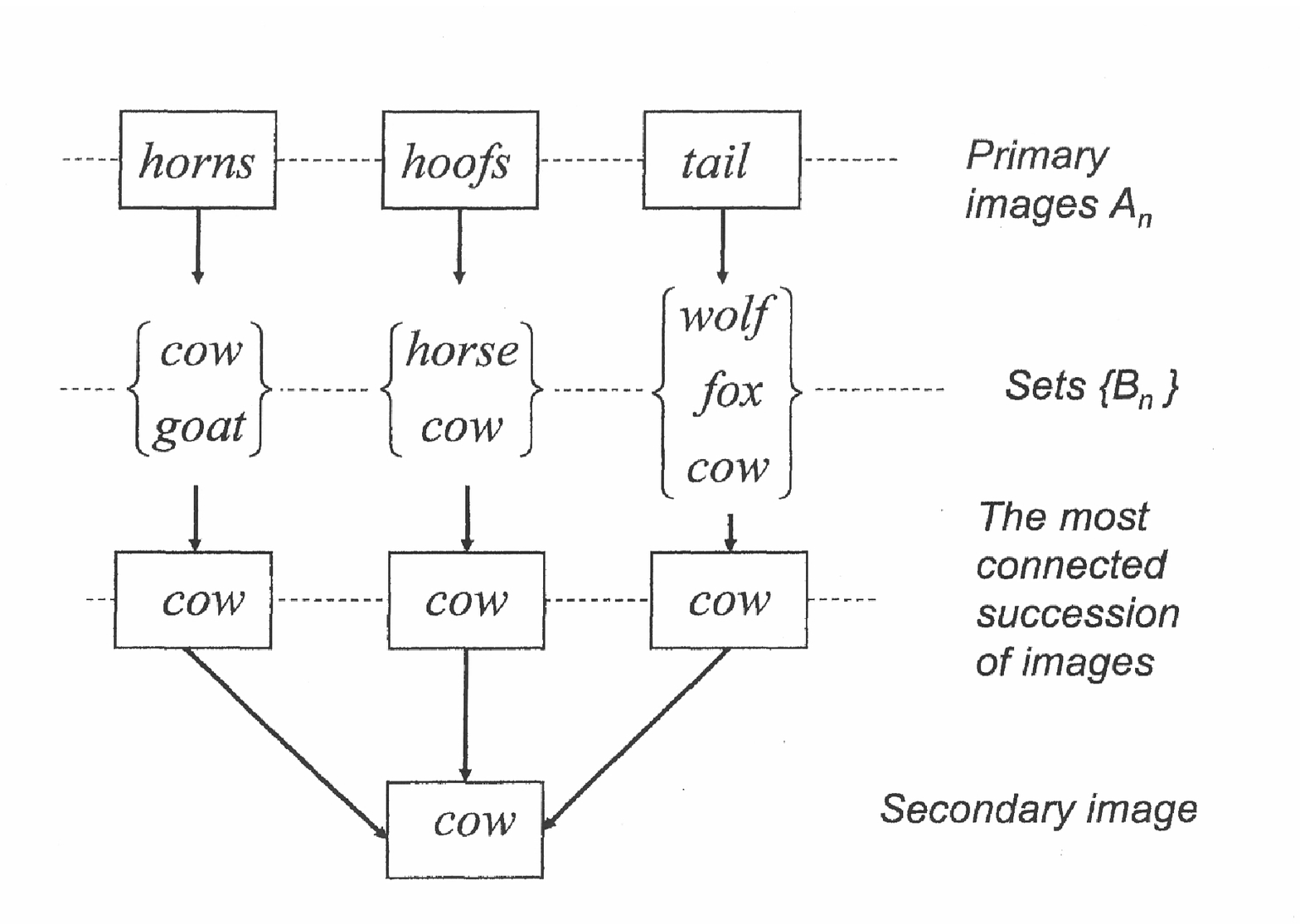}}
\caption{The scheme of enlarging the images. } \label{fig5}
\end{figure}
repetition of this image; after excluding repetition the latter takes
its place in the secondary succession (Fig.\,5).

Evidently, the specific malfunction described above can
take place in the process of developing the higher level versions,
even if the primary succession is determined uniquely: this
explains jokes with commutation of complex images (examples 5*,
6*, 7*, 8*, 9*, 11*).  The examples 6*, 7*, 8* shows that
emotional estimates constructed according to the described above
scheme can become one of the high level images.

\vspace{6mm}
\begin{center}
{\bf 9. The ambiguity scheme  }
\end{center}
\vspace{3mm}

Suppose that two versions $1$ and $2$ exist on the primary level;
on the second level the single version $1'$ corresponds to $1$
and the single version $2'$ corresponds to $2$ (Fig.\,6). Consider
\begin{figure}
\centerline{\includegraphics[width=5.1 in]{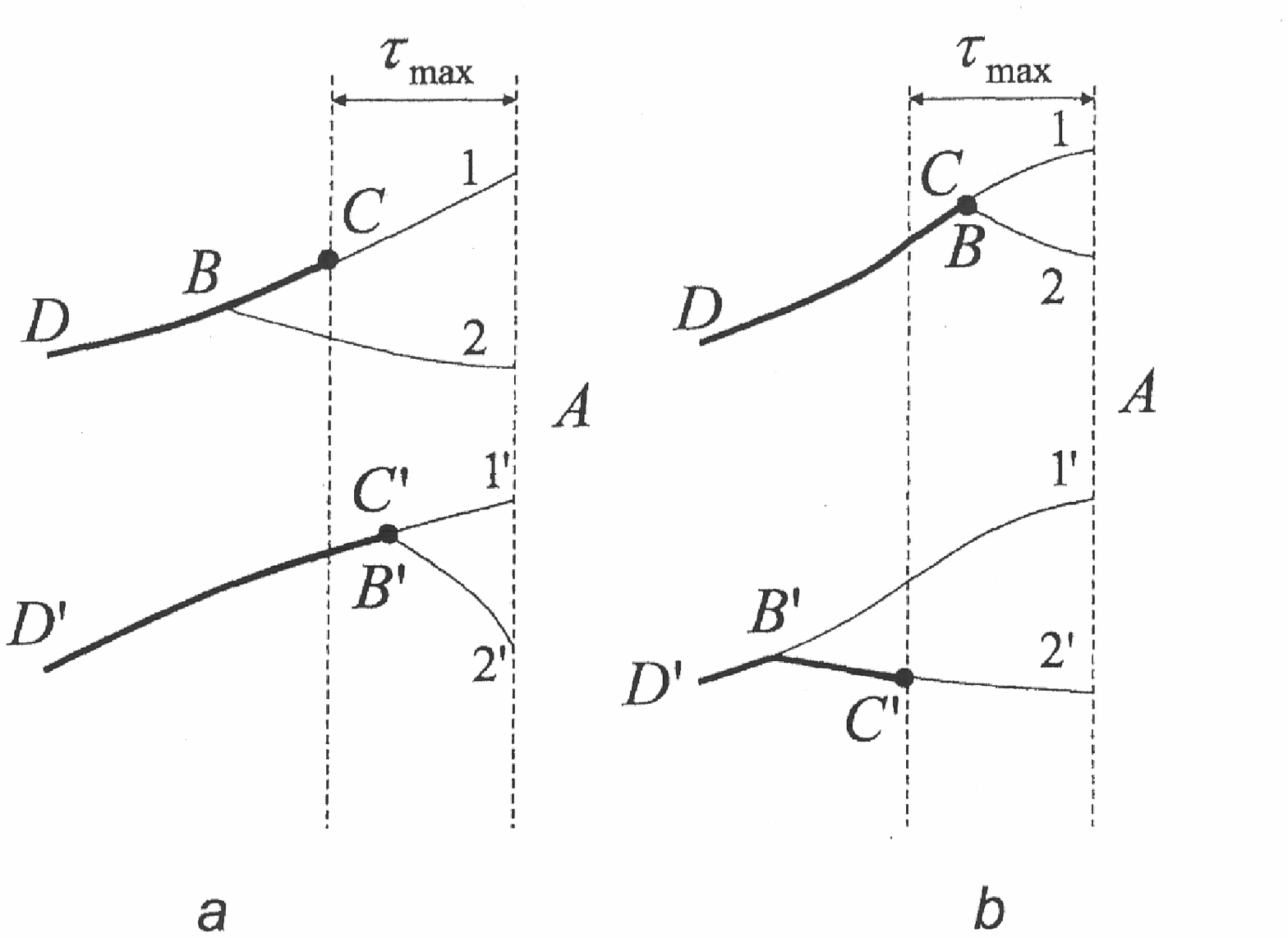}}
\caption{The situation corresponding to "the ambiguity
scheme": the order in which the trajectories are transmitted
into consciousness depends on the mutual arrangement of points
$B$ and $B'$. } \label{fig6}
\end{figure}
the case when version $1$ is more probable than $2$ and version
$2'$ is more probable than $1'$. Then "logically" it is
impossible to give preference to one of the two possibilities
$1+1'$ or $2+2'$:  each of them is "correct" in a certain sense.
Such situation takes place in  the process of the perception of
jokes constructed according to "the ambiguity scheme".

Indeed, on the primary level the connection between the images
nearest in succession is established (close  context) while
on the high levels the connection between remote images is
established (remote context). It is easy to see that in all
examples of Sec.\,2, corresponding to the ambiguity scheme,
one of the versions is more probable
in close  context,
another  --- in remote context:
\vspace{3mm}
$$
 \begin{array}{ccc}
{\rm Example}
& {\rm  Close\,\, context\,\,}
& {\rm Remote\,\, context} \\
{    } &\qquad\qquad {\rm  (or\,\, a\,\,separate\,\,word)}
\qquad \qquad { }\\
\hline
   &   &   \\
1^* & {\rm treated}\,\, familiarly     & {\rm
Rotshild}\, - \,millionaire\\
2^* & {\rm put\,\, in}\,\, reliable\,\, place
& confidence \,\,({\rm by \,\,the\,\, analogy})\\
3^* &{\rm her}\,\, virginity & guiltlessness\,\,{\rm of\,\,
Dreifus} \\
4^* & "Calcutta \,\,roast" & roast\,\, from\,\,
Calcutta \\
12^*  &
idealist  & {\rm complete}\,\, idiot \\
13^* & {\rm chair\,\, (for} \,\,sitting)
& {\rm chair\,\, (for}\,\, sleeping)
\end{array} \,,
$$

\vspace{3mm}
\noindent
So the first version corresponds to the variant
$1+1'$ while the second  --- to the variant $2+2'$.

 The uncertainty in the order of the appearance of the two
versions is connected with the unevenness of information
processing
in different channels, namely the mutual arrangement of
points $B$ and $B'$ with respect to front $A$ (Fig.\,6).
If point $B$ goes behind point $B'$ then version $1$ is
transmitted to the consciousness first (Fig.\,6,a); it takes
place up to the moment when delay  $\tau_{AB'}$ becomes
equil to $\tau_{max}$  ---  then version $2'$ is
transmtted to consciousness while version $1$ as incompatible
with $2'$ is deleted and replaced by $2$. If point $B$ outruns
point $B'$ then version $2'$ is transmitted to consciousness first
(Fig.\,6,b); when  $\tau_{AB}$ becomes equil to $\tau_{max}$
then the transmission of version $1$ begins, while version $2'$ is
replaced by $2$.
Formally, the considered situation is reduced to "the switching
scheme" but its psychologycal perception will be different.
Indeed, the relative order of following for points $B$ and $B'$
depends on the rates of processing in two channels, which
is determined by occasional reasons and can change for the
repeated reading of the text. The attempt to fix the order,
in which two versions arise, does not lead to a definite result
and this order appears to be elusive.

\vspace{6mm}
\begin{center}
{\bf 10. Perception of  the comic. }
\end{center}
\vspace{3mm}

A definition of the comic as "deviation from the norm" (Sec.\,2)
may arouse doubt, since not any deviation from the norm looks
funny.  However, one should have in mind two circumstances.
Firstly, the comic often give rather small humorous effect,
which can be easily forced out by side emotions (see Sec.\,6).
Since some background level of forcing out surely exists, so the
comic looks funny when a deviation from the norm exceeds a
certain critical level. Secondly, the habitual, oft-repeated
deviations do not produce the humorous effect by the same reason
as hackneyed
 jokes (Sec.\,7). Theoretically, any
deviation from the norm is comical, but practically it should be
sufficiently strong and unusual, in order to look funny.

The humorous effect from the comic arises by the same mechanism,
as for jokes based on the switching scheme (Sec.\,5, 8).
In spite to deviation from the norm, the comic still conserves
sufficient likelihood with the norm; so, at one of the higher
levels of information processing repeated switching of images
"the norm" --- "not the norm" occurs.

\vspace{6mm}
\begin{center}
{\bf 11. Development of a sense of humour in the process of
evolution }
\end{center}
\vspace{3mm}

Evolution of a sense of humour can be imagined in a following
manner. At the first stage of evolution
a sense of humour is surely absent,
since the algorithm of the information processing should be
sufficiently complicated for its existence: e.\,g. the condition
$M>1$ is necessary for the model of Sec.\,3. When the algorithm
becomes sufficiently complex and the specific malfuction
described in Sec.\,5 becomes possible, the second stage
of evolution begins:
a sense of humour exists, but  laughter (as a specific muscular
reaction) is still absent. The latter is related with the fact
that the main part of the neural network rules the muscle
performance and digestion,  while  thinking occurs in small
"islands" of intellect: dumping  of nervous energy
to the motor cortex occurs in occasional manner, and amount of
this energy is rather small. Thus, at this stage the humorous
effect is accompanied with only small muscular contractions and
is not noticed in the population. The growth of amount of the
intellectual activity results in the growth of the released
nervous energy, so muscular contractions becomes noticeable:
a necessity arises to arrange the discharge of energy,
in order to exclude movement of limbs and prevent the animal
from falling. As a result, the released nervous energy is
transmitted into specific muscular centers and laughter
arises as a characteristic muscular reaction to the humorous
effect.  The third step of evolution starts, when the humour is
recognized in the population and begins to play a social role.

Some of psychologists, and in particular Freud \cite{4},
consider the pleasure obtained from laughter as the main cause
of the existence of a sense of humour: a man discovers the
possibility of extracting pleasure from the psychical process and
begins subconsciously and then consciously to exploit it. Our
viewpoint is the opposite: a sense of humour is biologically
conditioned by the necessity to quicken the transmission of
information to consciousness and of a more effective use of brain
resources: so the pleasure obtained from laughter is not an
essential factor (similarly, the two reflexes --- sneezing and
coughing --- exist regardless of the pleasure afforded by the
first and the displeasure caused by the second, because they are
dictated by the biological necessity of cleaning out the
respiratory system).  Of course, if laughter afforded displeasure
the social function of humour would change: the society would try
to get rid of it by censorship, prosecution of witty people and
so on.

\vspace{6mm}
\begin{center}
{\bf 12. Indeed, can a computer laugh?  }
\end{center}
\vspace{3mm}

In the previous sections, we have demonstrated the principal
possibility to endow a computer by a sense of humour.
But is it really possible to create a computer program which will
"laugh" in the same cases as a man? From our viewpoint, it is
quite possible if we restrict ourselves to a primary level of
information processing, i.\,e. to the simplest types of
jokes based on switching of meanings of individual words
(example 14*). The corresponding program will have approximately
the same level of comlexity, as the average program for
machine translation \cite{12,13,14} and its creation is possible
as a result of work of the whole group of researches during
several years. However, it should be noted that such activity is
a natural next step in the machine translation researches: in the
case of ambiguity, the existing programs suggest several
variants for a choice of consumer, while the proper
treatment of ambiguity is crucial for registration
of the humorous effect. Development of "a computer with humour"
should be naturally based on existing programs for machine
translation.

As for teaching  a computer to react on the more complex
samples of humour related with the higher levels of information
processing, it looks incredible at the present time. In order to
do it, one should reveal a complete set of images the average
human brain contains and to establish the correct associative
links between these images.  This would require many years of
work of psychologists and programmers.

\end{document}